\documentclass{article}

\usepackage{arxiv}

\usepackage[utf8]{inputenc} % allow utf-8 input
\usepackage[T1]{fontenc}    % use 8-bit T1 fonts
\usepackage[authoryear,round]{natbib}

\usepackage[dvipsnames]{xcolor}
\definecolor{citegreen}{RGB}{34,120,34}

\usepackage[colorlinks=true,
            citecolor=citegreen,
            linkcolor=black,
            urlcolor=blue]{hyperref}
\usepackage{url}            % simple URL typesetting
\usepackage{booktabs}       % professional-quality tables
\usepackage{amsfonts}       % blackboard math symbols
\usepackage{nicefrac}       % compact symbols for 1/2, etc.
\usepackage{microtype}      % microtypography
\usepackage{graphicx}
\usepackage{doi}
\usepackage{amsmath}
\usepackage{float}
\usepackage{subcaption}

\usepackage{multirow}
\usepackage{makecell}
\newcommand{\imp}{\textcolor{ForestGreen}{\scriptsize$\downarrow$}}
\newcommand{\degd}{\textcolor{BrickRed}{\scriptsize$\uparrow$}}

\newcommand{\IntroText}{
\section{Introduction}
Large language models (LLMs) are increasingly deployed for reasoning and decision support in high-stakes domains such as medicine, finance, and ecology~\citep{singhal2023large, li2023large, mora2024uncertainty}, where reliable uncertainty quantification (UQ) is as important as predictive accuracy. Yet estimating confidence for black-box LLMs, accessible only through APIs, remains an open problem. The problem is further complicated in hierarchical reasoning, where outputs span multiple semantic levels and abstention can yield partial outputs rather than a single flat label.

Biodiversity monitoring provides a particularly challenging instance of this setting. Taxonomic predictions are hierarchical and strongly long-tailed~\citep{gharaee2023step}. Rare and endangered species are difficult to monitor and can play disproportionately important roles in ecosystems~\citep{dee2019ecosystem}, making UQ essential for routing uncertain cases to domain experts. A recent system~\citep{lesperance2025taxonomic} combines vision language model (VLM)-based image captioning, retrieval-augmented generation (RAG), and LLM-based reasoning to generate interpretable predictions across taxonomic ranks. However, it lacks numerical confidence estimates, relying instead on qualitative abstentions that do not support principled rejection thresholds.

We address this gap by studying uncertainty estimation for black-box LLMs in hierarchical taxonomic reasoning, where only externally accessible signals are available. Our contributions are threefold:
\begin{enumerate}
    \item \textbf{Formulation.} We cast black-box LLM taxonomic UQ as rank-level selective prediction under partial-path abstention.

    \item \textbf{Method.} We develop lightweight supervised uncertainty estimators that use proxy-LLM features, hierarchy-aware supervision, and rank-aware output parameterizations to predict rank-wise correctness.

    \item \textbf{Design insight.} On a realistic long-tailed arthropod pipeline, supervised estimators outperform a token-likelihood baseline for global-threshold selective prediction, and a rank-specific multi-head design performs best overall.

\end{enumerate}
}

\newcommand{\RelatedWorkText}{
\section{Related Work}
\textbf{Uncertainty estimation (UE) for LLMs.}
LLM uncertainty estimation has been studied in both white-box and black-box settings. White-box methods leverage logits or hidden states~\citep{geng-etal-2024-survey}, but these signals are unavailable for proprietary API models. Black-box approaches typically rely on self-verbalized confidence~\citep{tian2023just} or consistency or semantic clustering across multiple sampled responses~\citep{farquhar2024detecting, manakul2023selfcheckgpt}. The former may fail to faithfully reflect the model's intrinsic uncertainty~\citep{yona-etal-2024-large}, while the latter can be prohibitively expensive in API-based pipelines. We therefore build on the supervised black-box approach of Liu \textit{et al.}~\citep{liu2024uncertainty}, which uses an open-source \emph{tool LLM} to extract proxy features from prompt-response pairs and trains a correctness predictor for the target model. We view tool-LLM features as a practical proxy signal rather than faithful reconstruct of target uncertainty.

\textbf{Hierarchical reasoning in taxonomic prediction.}
Taxonomic reasoning is a challenging setting for uncertainty estimation because predictions are hierarchical, long-tailed, and often partially specified through abstention. Prior work on hierarchical classification has developed evaluation measures based on ancestor overlap and partial credit~\citep{kosmopoulos2015evaluation}, and recent work has extended this perspective to taxonomy-aware evaluation~\citep{snaebjarnarson2025taxonomy}. In parallel, an LLM-based pipeline has been proposed for taxonomic reasoning in rare arthropods~\citep{lesperance2025taxonomic}. Our focus is complementary. Rather than improving the upstream reasoning pipeline, we study UQ for its hierarchical outputs and use hierarchy-aware signals as auxiliary supervision.
}

\newcommand{\MethodologyText}{
\section{Methodology}
We formulate uncertainty estimation as a supervised prediction problem (Figure~\ref{fig:workflow}).

\begin{figure}[th]
    \centering
    \includegraphics[trim=0 14.6cm 0 0cm, clip, width=0.8\textwidth]{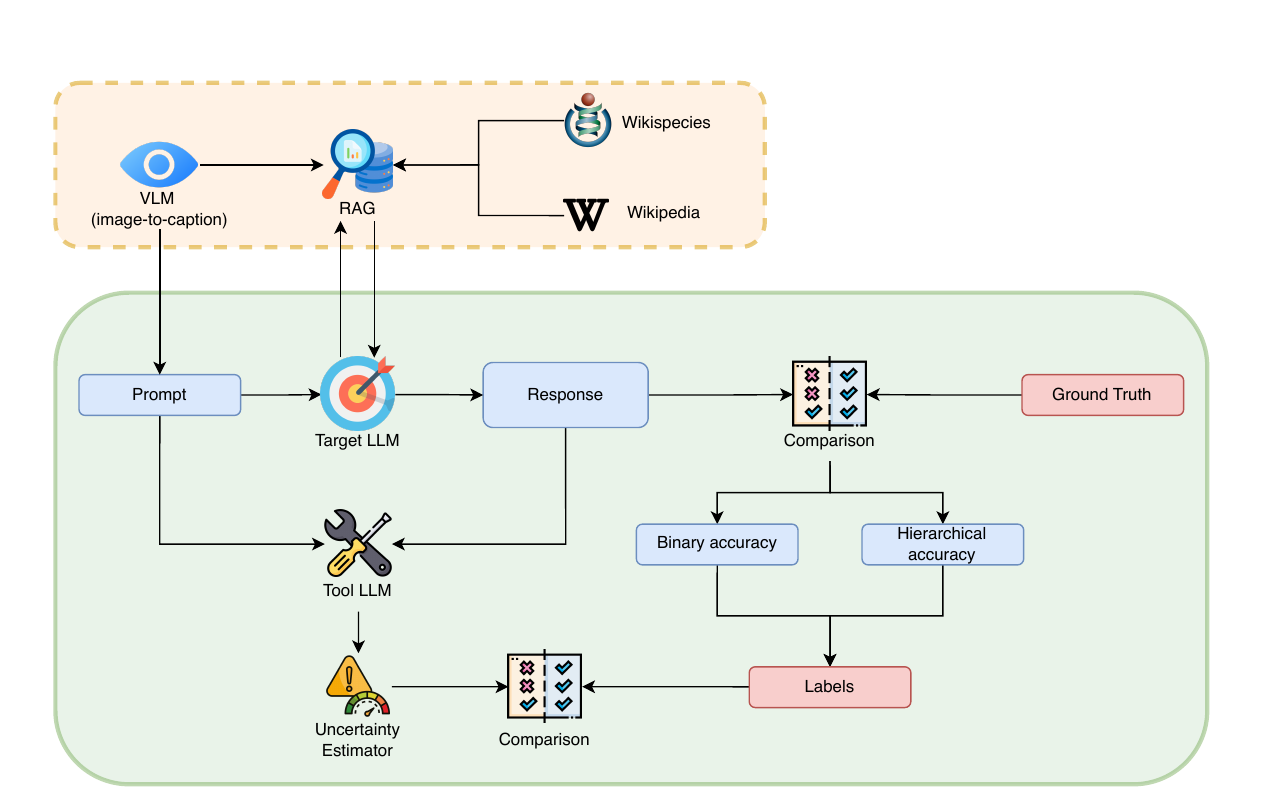}
    \caption{End-to-end pipeline for LLM uncertainty estimation. A target LLM generates taxonomic predictions from VLM captions and RAG-retrieved context. A tool LLM extracts proxy features from each prompt–response pair, which a lightweight uncertainty estimator maps to a confidence score. Blue boxes denote inputs and intermediate outputs. Red boxes denote prediction targets.}
    \label{fig:workflow}
\end{figure}

\subsection{Dataset and Upstream Taxonomic Reasoning System}
This study uses 829 arthropod images from the Rare Species dataset~\citep{stevens2024bioclip}, obtained after filtering invalid samples from an initial set of 951. Each sample is associated with a complete taxonomic path, spanning seven taxonomic ranks. The dataset is conservation-biased, with all species belonging to International Union for Conservation of Nature (IUCN) Red List categories from Near Threatened to Extinct in the Wild.

As input to our uncertainty estimator, we use the textual taxonomic outputs from the upstream simple-RAG reasoning system~\citep{lesperance2025taxonomic}, instantiated with GPT-4o~\citep{hurst2024gpt} as the target LLM. The system predicts taxa in a coarse-to-fine manner and abstains at the first uncertain rank, producing variable-depth partial paths and consequently limited coverage at finer ranks (Appendix Table~\ref{tab:upstream_results}). We therefore model uncertainty at the rank level to capture both error propagation and partial-path abstention.

\subsection{Supervision Labels}
Each rank-level instance has two supervision signals:(i)~\emph{a binary accuracy} $y \in \{0,1\}$, indicating whether the predicted taxon matches the ground truth at that rank; and (ii)~\emph{a hierarchical accuracy} $h \in [0,1]$, which assigns partial credit based on taxonomic overlap. Following~\citep{snaebjarnarson2025taxonomy}, we compute hierarchical precision, recall, and F1. For predicted node $v^{\text{pred}}$ and gold node $v^{\text{gold}}$, letting $\mathrm{anc}(v)$ denote the set of ancestors of $v$, including $v$ itself,

\begin{equation}
hP = \frac{|\mathrm{anc}(v^{\text{pred}}) \cap \mathrm{anc}(v^{\text{gold}})|}
           {|\mathrm{anc}(v^{\text{pred}})|}, \quad
hR = \frac{|\mathrm{anc}(v^{\text{pred}}) \cap \mathrm{anc}(v^{\text{gold}})|}
           {|\mathrm{anc}(v^{\text{gold}})|}, \quad
hF = \frac{2 hP hR}{hP + hR}.
\label{eq:hierarchical_equations}
\end{equation}
We set $h = hF$ and use it only as an auxiliary training target. Kingdom and phylum are excluded, as they are constant across all samples and hence non-discriminative.

\subsection{Tool LLM Feature Extraction}\label{sec:feature-extraction}
Following Liu \textit{et al.}~\citep{liu2024uncertainty}, we use an open-source tool LLM to extract proxy features from each prompt-response pair. We consider Gemma 7B~\citep{team2024gemma}, GPT-OSS 20B~\citep{agarwal2025gpt}, or Qwen3 30B A3B Instruct~\citep{qwen3technicalreport} as tool LLMs (layer details in Appendix Table~\ref{tab:tool_llms}). We collect two feature families (Appendix Figure~\ref{fig:feature-extraction_feature-tree}): (i)~\emph{representation features}, namely averaged and last-token hidden states from the middle layer of the tool LLM, and (ii)~\emph{distributional features}, including statistics of entropy, negative log-probability, and raw token probability.

\subsection{Supervised Uncertainty Estimators}
To mitigate overfitting under sparse fine-rank supervision, we use a compact MLP with a shared two-layer trunk (hidden size 16, ReLU, dropout 0.3) and task-specific head(s). The primary head is a sigmoid classifier that predicts the probability of rank-wise correctness, directly supporting AUROC and risk--coverage evaluation. Some variants add a regression head for hierarchical F1 to provide hierarchy-aware auxiliary supervision.

We train the primary head with binary cross-entropy (BCE) and the auxiliary head with mean squared error (MSE):
\begin{equation}
\mathcal{L}
=
\underbrace{\mathrm{BCE}(\hat{p}, y)}_{\mathcal{L}_{\mathrm{BCE}}}
+
\underbrace{\lambda \lVert \hat{h} - h \rVert_2^2}_{\mathcal{L}_{\mathrm{MSE}}}.
\label{eq:loss_function}
\end{equation}
Here, $\hat{p}$ denotes the predicted probability of rank-wise correctness, $y$ the binary correctness label, $\hat{h}$ the predicted hierarchical score, and $h$ the hierarchical-F1 auxiliary target.

We study three variants: \textbf{H1}, a shared MLP trunk with a single binary head trained with $\mathcal{L}_{\mathrm{BCE}}$ only; \textbf{H2}, the same trunk augmented with an auxiliary regression head and trained with Eq.~\eqref{eq:loss_function}; and \textbf{H3}, a shared trunk with five rank-specific binary heads plus one auxiliary regression head. BCE is applied only to the head matching the sample rank, while the auxiliary regression loss is applied globally. At inference, H3 produces correctness probabilities for all five ranks in a single forward pass. Implementation details are provided in Appendix~\ref{app:implementation_details}.\looseness-1

\subsection{Negative log-likelihood (NLL) baseline}
As a training-free baseline, we use the negative mean token NLL of the target LLM output~\citep{lin2024contextualized, gupta2024language}. For generated output $\mathbf{y}=(y_1,\ldots,y_T)$ under context $\mathbf{c}$, we define
\begin{equation}
\label{eq:mean_nll}
\mathrm{conf}_{\mathrm{nll}}(\mathbf{y}\mid \mathbf{c})=-\mathrm{NLL}_{\mathrm{mean}}(\mathbf{y}\mid \mathbf{c})
= \frac{1}{T}\sum_{t=1}^{T}\log p_{\theta}\!\left(y_t \mid \mathbf{c}, y_{<t}\right),
\end{equation}
where $p_{\theta}(y_t \mid \mathbf{c}, y_{<t})$ denotes the token probability returned by the target LLM, and $T=|\mathbf{y}|$ is the output length in tokens. We use this baseline because it is available in a single GPT-4o API pass, whereas full-distribution or multi-sample baselines are unavailable or costly.

\subsection{Evaluation protocol}
We evaluate uncertainty at the rank level. Each image--rank pair $(x,r)$ is one example, with correctness label 
$
y^{\mathrm{corr}}_{r}(x)=\mathbf{1}\!\left[\hat{y}^{\mathrm{tax}}_{r}(x)=y^{\mathrm{tax}}_{r}(x)\right]
$ 
and confidence score $\mathrm{conf}(x,r)$. 

\textbf{Risk--coverage Curve.}
To reflect deployment under a single global rejection rule, we pool validation examples across ranks and, for a target coverage $c$, choose a threshold $\tau=\tau(c)$ as the $(1-c)$-quantile of confidence scores. On the test set, we accept examples with $\mathrm{conf}(x,r)\ge \tau$, and define selective risk as the error rate among accepted examples~\citep{geifman2019selectivenet}:

\begin{equation}
R(\tau)=
\frac{
\mathbb{E}\!\left[
\bigl(1-y^{\mathrm{corr}}_{r}(x)\bigr)\,
\mathbf{1}\!\left[\mathrm{conf}(x,r)\ge \tau\right]
\right]
}{
\mathbb{E}\!\left[
\mathbf{1}\!\left[\mathrm{conf}(x,r)\ge \tau\right]
\right]
}.
\end{equation}

\textbf{AUROC.} 
We report micro AUROC (pooled across ranks) and macro AUROC
(per-rank average) as threshold-free discrimination metrics,
with $y^{\mathrm{corr}}_{r}(x)=1$ as the positive class.

}

\newcommand{\ResultsText}{
\section{Results}
\begin{table}[t]
\centering
\caption{All-rank micro AUROC (mean $\pm$ std over 5 seeds). Highest mean is bolded.}
\label{tab:all-rank_aggregated_micro_auroc}
\resizebox{0.9\textwidth}{!}{
\begin{tabular}{lcccc}
\toprule
Tool LLM & \multicolumn{1}{c}{Benchmarks} & \multicolumn{3}{c}{Ours} \\
\cmidrule(lr){2-2}\cmidrule(lr){3-5}
 & Baseline & H1 & H2 & H3 \\
\midrule
Gemma 7B &\multirow{3}{*}{0.573} & 0.710 $\pm$ 0.050 & 0.694 $\pm$ 0.060 & \textbf{0.796 $\pm$ 0.023} \\
Qwen3 30B A3B Instruct& & 0.743 $\pm$ 0.050 & 0.719 $\pm$ 0.080 & \textbf{0.749 $\pm$ 0.017} \\
GPT-OSS 20B & & 0.733 $\pm$ 0.069 & 0.783 $\pm$ 0.044 & \textbf{0.790 $\pm$ 0.028} \\
\bottomrule
\end{tabular}}
\end{table}

\begin{figure}[t]
    \centering
    \begin{subfigure}[t]{0.32\textwidth}
        \centering
        \includegraphics[width=\textwidth]{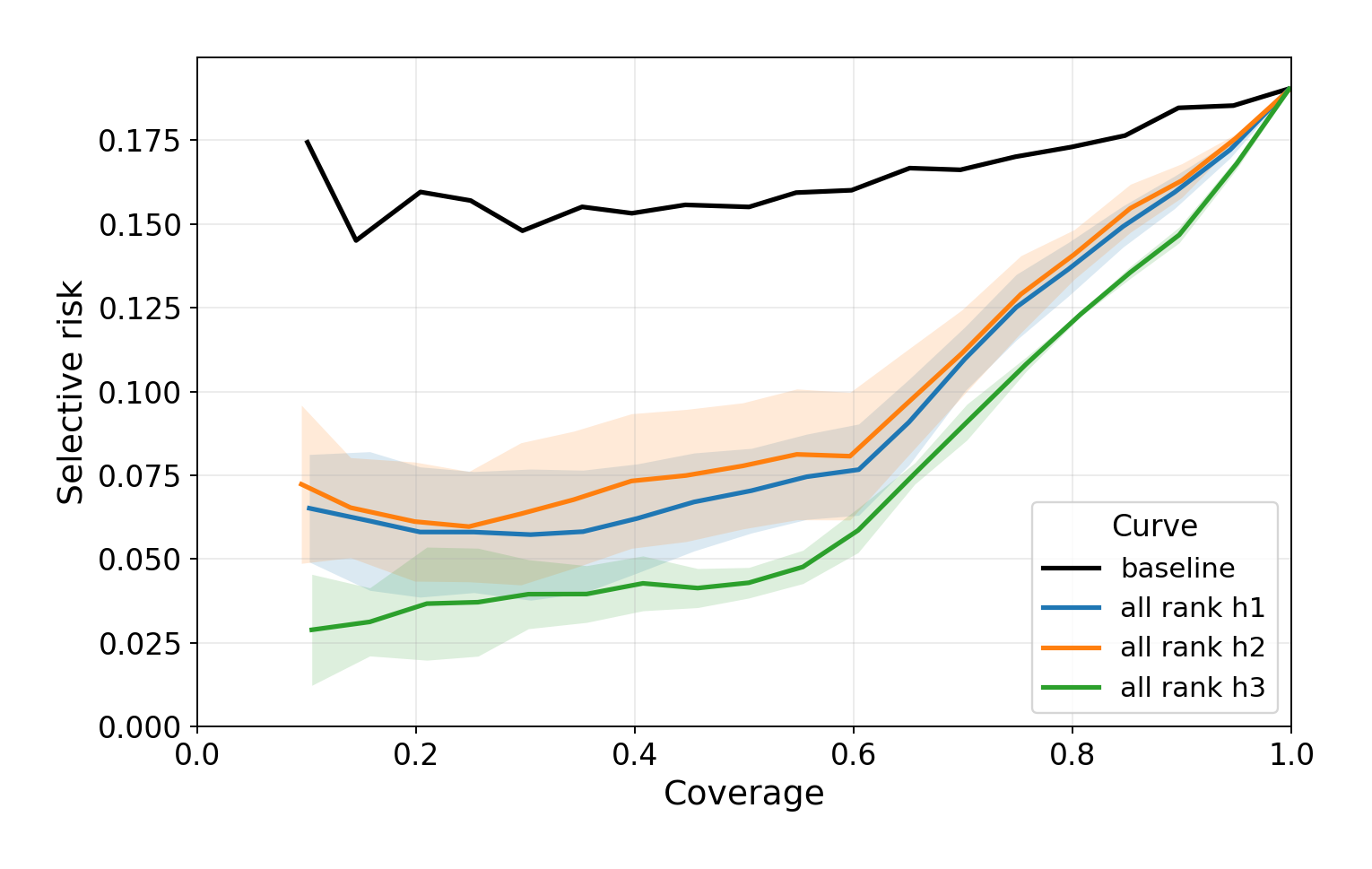}
        \caption{Gemma 7B.}
        \label{fig:all_rank_aggregated_selective_risk_gemma}
    \end{subfigure}
        \begin{subfigure}[t]{0.32\textwidth}
        \centering
        \includegraphics[width=\textwidth]{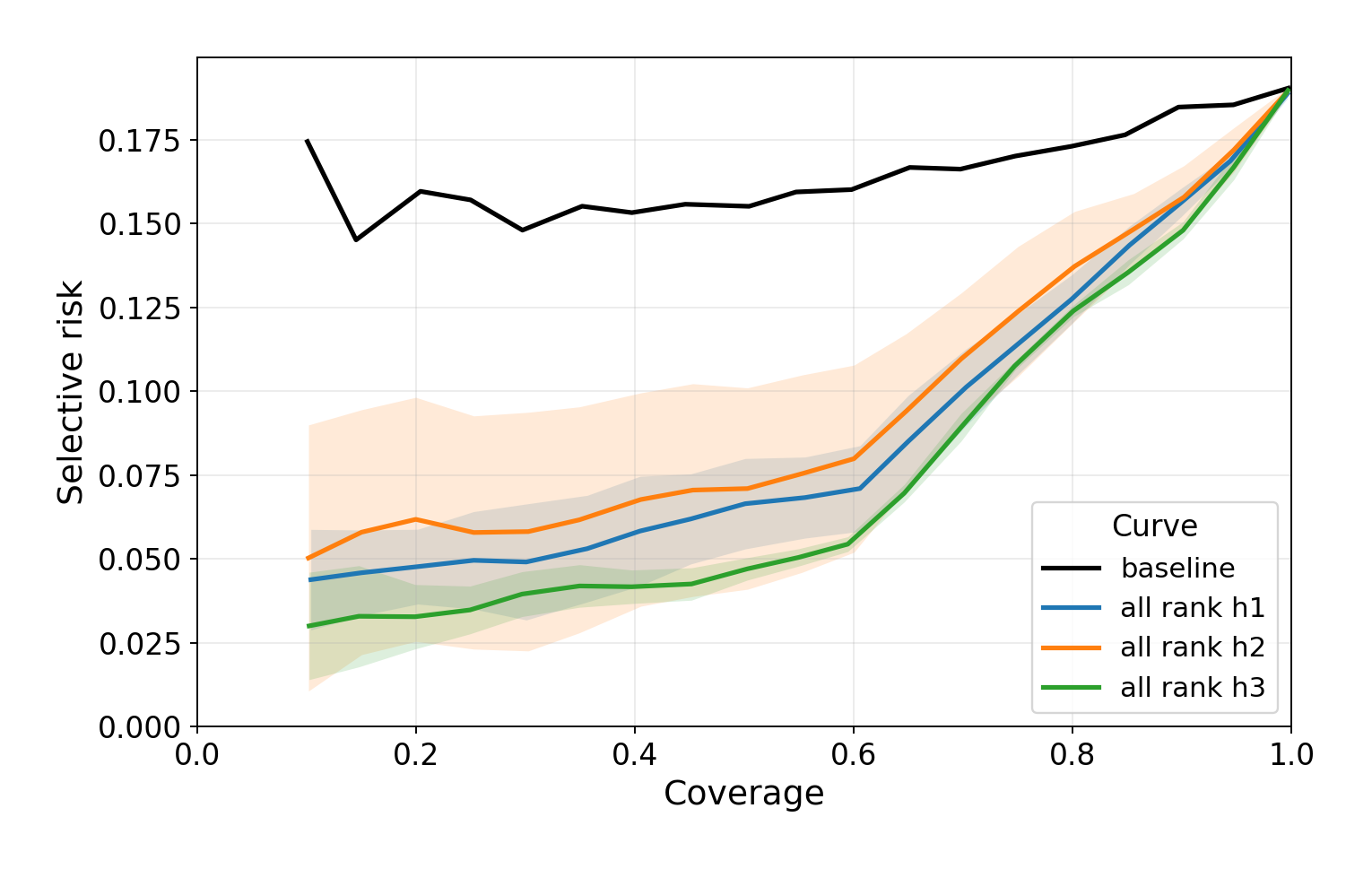}
        \caption{Qwen3 30B A3B Instruct.}
        \label{fig:all_rank_aggregated_selective_risk_qwen3}
    \end{subfigure}
    \begin{subfigure}[t]{0.32\textwidth}
        \centering
        \includegraphics[width=\textwidth]{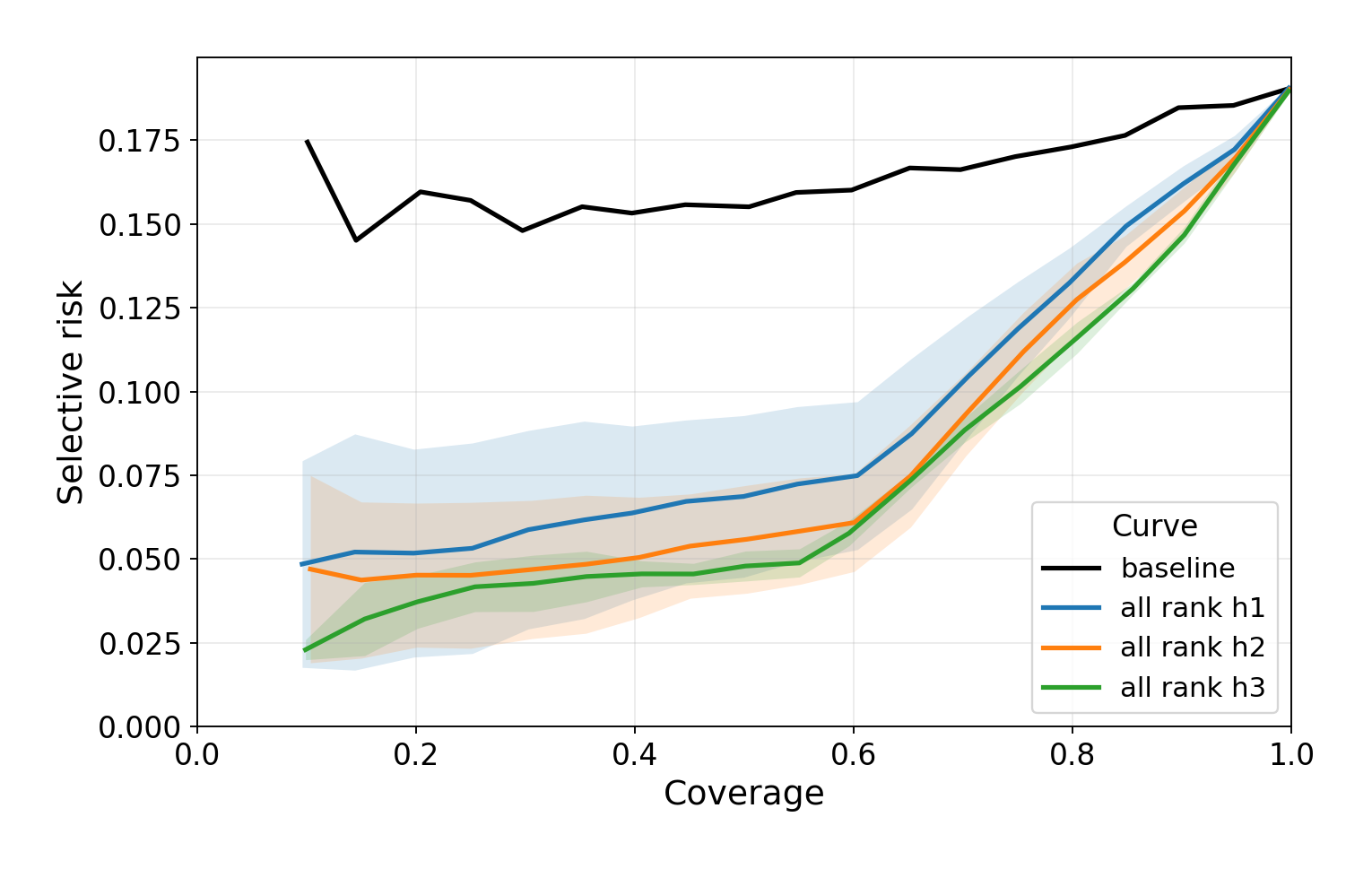}
        \caption{GPT-OSS 20B.}
        \label{fig:all_rank_aggregated_selective_risk_gpt}
    \end{subfigure}
    \caption{Global-threshold risk--coverage curve. Lower is better. Curves show the mean over 5 random seeds, with shaded regions indicating $\pm 1$ standard deviation.}
    \label{fig:all_rank_aggregated_selective_risk_all}
\end{figure}

Table~\ref{tab:all-rank_aggregated_micro_auroc} shows that supervised uncertainty estimators consistently and substantially outperform the mean-NLL baseline in micro AUROC across all three tool LLMs. H3 achieves the highest mean micro AUROC overall, suggesting the strongest discrimination between correct and incorrect, although its margin over H1 is small for Qwen3 and clearer for Gemma 7B and GPT-OSS 20B.

Figure~\ref{fig:all_rank_aggregated_selective_risk_all} provides the most deployment-relevant comparison under a single global rejection threshold. Across all three tool LLMs, our supervised uncertainty estimators achieve remarkably lower selective risk than the mean-NLL baseline over a broad coverage range, with the largest gains at low-to-mid coverage. H3 consistently yields the best overall risk–coverage trade-off, while H1 and H2 also remain better than the baseline throughout.

The gains are less consistent under rank-balanced macro AUROC metric (Appendix Table~\ref{tab:all-rank_aggregated_macro_auroc} and Figure~\ref{fig:per_rank_cleveland_dot_pre_calibrated_all}). Calibration results are also reported in Appendix Table~\ref{tab:all_rank_calibration_metrics_all_models} and generally favour the supervised estimators over the mean-NLL baseline.
}

\newcommand{\DiscussionText}{
\section{Discussion}
The divergence between micro (Table~\ref{tab:all-rank_aggregated_micro_auroc}) and macro AUROC (Appendix Table~\ref{tab:all-rank_aggregated_macro_auroc}) stems from rank imbalance. Micro AUROC is dominated by high-coverage coarse ranks, whereas macro AUROC weights each rank equally. Per-rank results (Appendix Figure~\ref{fig:per_rank_cleveland_dot_pre_calibrated_all}) suggest that the degradation is concentrated at deeper ranks, where frequent abstentions reduce the available training signal and increase variance (Appendix Table~\ref{tab:upstream_results}). Therefore, the rank-balanced metric exposes weaknesses that the deployment-oriented micro metric downweights.

For unified-threshold deployment, cross-rank comparability is key. H2 yields modest gains over H1 for most tool LLMs, except with GPT-OSS 20B, while H3 performs best overall (Figure~\ref{fig:all_rank_aggregated_selective_risk_all}). Since H2 and H3 use the same auxiliary supervision, this pattern suggests that rank-specific output heads matter more than the auxiliary hierarchical target alone.

A main limitation is sparse fine-rank supervision. Our study is further restricted to a single taxonomic reasoning pipeline and a global abstention threshold chosen to reflect deployment constraints. Future work should assess whether these conclusions generalize to richer fine-rank coverage,
wider distribution shifts, and rank-specific decision rules.
}

\newcommand{\ConclusionText}{
\section{Conclusion}
We studied UQ for black-box LLMs in hierarchical taxonomic reasoning under long-tailed data and partial-path abstention. Using proxy features extracted by a tool LLM, we trained lightweight supervised estimators with hierarchy-aware signals to predict rank-wise correctness. Across tool LLMs, these estimators outperformed a NLL baseline in micro discrimination and global-threshold selective prediction. A rank-specific multi-head design performed best overall, suggesting that estimator design for hierarchical outputs is as important as the uncertainty signal itself in unified-threshold deployment.
}

\renewcommand{\undertitle}{}
\renewcommand{\headeright}{}
\date{}

\title{Hierarchy-Aware Supervised Uncertainty Estimation for Black-box LLM Taxonomic Reasoning}

\author{
Shuting Xie$^{1}$,
Nathaniel Lesperance$^{2,1}$,
Graham W. Taylor$^{2,1,*}$\\
$^{1}$Vector Institute for AI, Toronto, ON\\
$^{2}$University of Guelph, Guelph, ON\\
$^{*}$Corresponding author: \texttt{gwtaylor@uoguelph.ca}
}

\renewcommand{\shorttitle}{Hierarchy-Aware Uncertainty Estimation}

\hypersetup{
pdftitle={Hierarchy-Aware Supervised Uncertainty Estimation for Black-box LLM Taxonomic Reasoning},
pdfsubject={},
pdfauthor={Shuting Xie, Nathaniel Lesperance, Graham W. Taylor},
pdfkeywords={Uncertainty, Large Language Model, Hierarchical Classification},
}

\begin{document}
\begin{center}
\small Accepted for publication in the Proceedings of the 39th Canadian Conference on Artificial Intelligence (Canadian AI 2026)
\end{center}

\vspace{1.5cm}

\maketitle

\begin{abstract}
Large language models (LLMs) are increasingly used for scientific decision support, yet reliable confidence estimation remains difficult in black-box settings. We study uncertainty estimation for hierarchical taxonomic reasoning generated by a black-box LLM in a long-tailed biodiversity monitoring pipeline. Using proxy features extracted by an open-source tool LLM, we train lightweight supervised estimators with hierarchy-aware supervision to predict rank-wise correctness. Across three tool LLMs, the supervised estimators consistently outperform a token-likelihood baseline for micro discrimination and selective prediction under a single global rejection threshold, improving micro AUROC from 0.57 to 0.75--0.80. The best results are achieved by a rank-specific multi-head design (H3), suggesting that accounting for hierarchical output structure is important when a unified abstention rule is required. Our code is publicly available at \url{https://github.com/uoguelph-mlrg/hierarchy-aware-llm-uq}
\end{abstract}

% keywords can be removed

\IntroText

\RelatedWorkText

\MethodologyText

\ResultsText

\DiscussionText

\ConclusionText

\section{Acknowledgments}
Resources used in preparing this research were provided, in part, by the Province of Ontario, the Government of Canada through CIFAR, and companies sponsoring the Vector Institute \url{http://www.vectorinstitute.ai/#partners}

\bibliographystyle{unsrtnat}
\bibliography{references}

\newpage
\appendix
\section{Implementation details}
\label{app:implementation_details}
All MLPs are trained with Adam (lr $=10^{-3}$, weight decay $=3\times10^{-4}$), batch size 128, for up to 200 epochs. We use early stopping on validation micro AUROC (patience $=5$, min $\Delta=10^{-4}$). For hierarchical supervision we set $\lambda=0.1$. An $\ell_1$ penalty ($10^{-4}$) is applied to the first linear layer weights (input-to-hidden) to encourage sparse use of input features. Feature selection is applied only to the hidden-state representation features, combining the top-100 features from each of Lasso regression, mutual information, and Pearson correlation with the target. We use 5-fold cross-validation. To avoid specimen leakage, all rank-level records derived from the same specimen are assigned to the same cross-validation fold.

% ---------------------------------------------------------
\section{Upstream taxonomic pipeline statistics}
\label{app:upstream_pipeline}

\begin{table}[H]
\centering
% \small
% \setlength{\tabcolsep}{5pt}
\caption{Rank coverage and prediction accuracy in the upstream system before filtering.}
\label{tab:upstream_results}
\begin{tabular}{lrrcccccc}
\toprule
\textbf{Rank} & \textbf{Attempts} & \textbf{Coverage (\%)} & \textbf{Accuracy} & \textbf{F1} & \textbf{HP} & \textbf{HR} & \textbf{HF} \\
\midrule
Class   & 950 & 100 & 0.966 & 0.973 & 0.981 & 0.981 & 0.981 \\
Order   & 922 & 97  & 0.872 & 0.885 & 0.955 & 0.955 & 0.955 \\
Family  & 745 & 78  & 0.693 & 0.710 & 0.914 & 0.914 & 0.914 \\
Genus   & 223 & 23  & 0.453 & 0.532 & 0.865 & 0.865 & 0.865 \\
Species & 55  & 6   & 0.455 & 0.500 & 0.847 & 0.847 & 0.847 \\
\bottomrule
\end{tabular}
\end{table}

% ---------------------------------------------------------
\section{Tool LLM configurations and extracted features}
\label{app:tool_llms_and_features}

\begin{table}[H]
% \small
\caption{Details of tool LLMs. The first column reports the model name, followed by the exact checkpoint identifier in parentheses.}
\label{tab:tool_llms}
\centering
\begin{tabular}{lccc}
\toprule
Tool LLM & Middle layer & Total Number of Layers & Hidden size \\
\midrule
\makecell[l]{Gemma 7B \\ \small(\texttt{google/gemma-7b})} & 14 & 28 & 3072 \\
\makecell[l]{Qwen3 30B A3B Instruct \\ \small(\texttt{Qwen/Qwen3-30B-A3B-Instruct-2507-FP8})} & 24 & 48 & 2048 \\
\makecell[l]{GPT-OSS 20B \\ \small(\texttt{openai/gpt-oss-20b})} & 12 & 24 & 2880 \\
\bottomrule
\end{tabular}
\end{table}

% ---------------------------------------------------------
\begin{figure}[H]
    \centering
    \includegraphics[trim=0 20cm 0 0, clip, width=\textwidth]{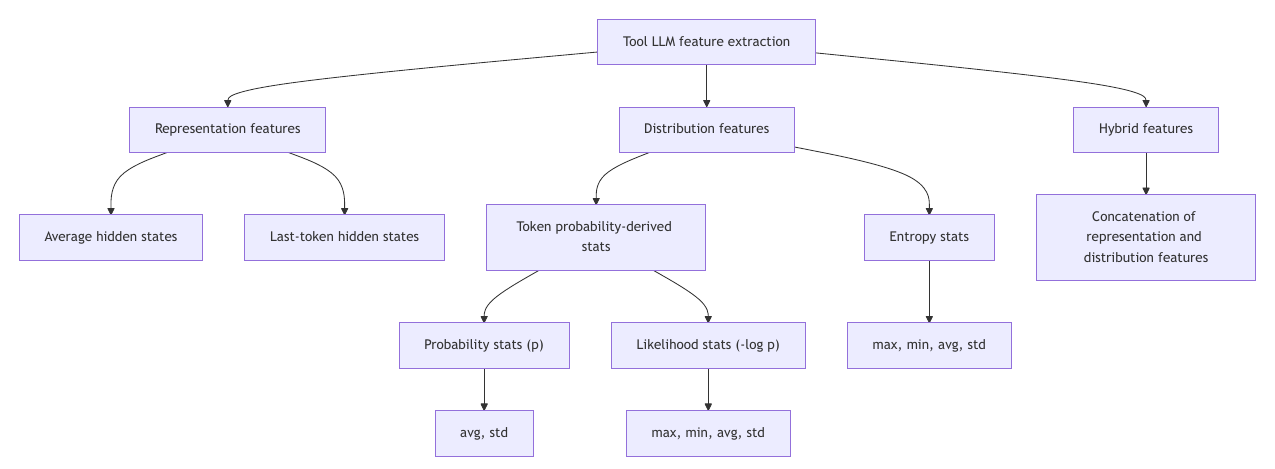}
    \caption{Feature tree demonstrates the feature types extracted by tool LLMs.}
    \label{fig:feature-extraction_feature-tree}
\end{figure}

% ---------------------------------------------------------
\section{Additional UQ results}
\label{app:additional_uq_results}

\begin{table}[H]
\centering
% \small
\caption{All-rank aggregated macro AUROC across tool LLMs. Results are reported as mean ± std over 5 random seeds. Bold indicates the best macro AUROC across all methods, and underlining marks the best performance among our three uncertainty estimators.}
\label{tab:all-rank_aggregated_macro_auroc}
\begin{tabular}{lcccc}
\toprule
Tool LLM & \multicolumn{1}{c}{Benchmarks} & \multicolumn{3}{c}{Ours} \\
\cmidrule(lr){2-2}\cmidrule(lr){3-5}
 & Baseline & H1 & H2 & H3 \\
\midrule
Gemma 7B & \multirow{3}{*}{\textbf{0.621}} & 0.451 $\pm$ 0.009 & 0.454 $\pm$ 0.013 & \underline{0.498 $\pm$ 0.036} \\
Qwen3 30B A3B Instruct & & \underline{0.520 $\pm$ 0.015} & 0.514 $\pm$ 0.016 & 0.491 $\pm$ 0.021 \\
GPT-OSS 20B & & 0.509 $\pm$ 0.029 & \underline{0.528 $\pm$ 0.044} & 0.515 $\pm$ 0.031 \\
\bottomrule
\end{tabular}
\end{table}

% ---------------------------------------------------------
\begin{figure}[H]
    \centering
    \begin{subfigure}[t]{0.32\textwidth}
        \centering
        \includegraphics[width=\textwidth]{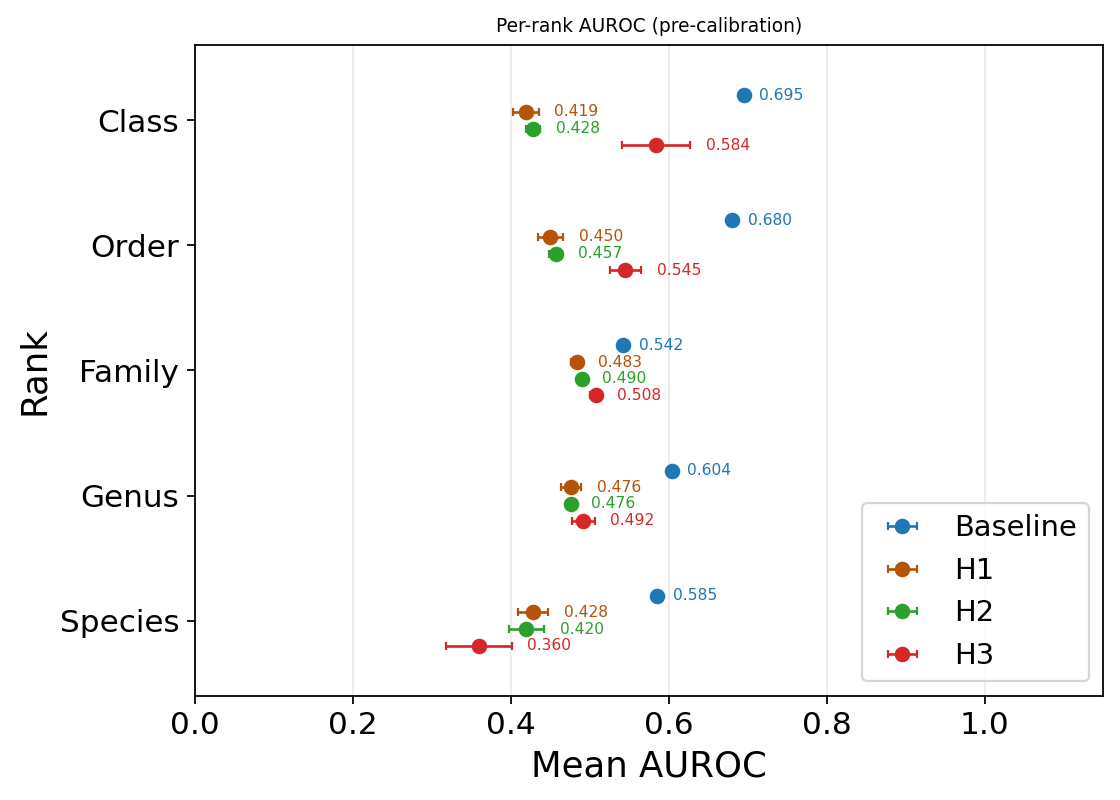}
        \caption{Gemma 7B.}
        \label{fig:per_rank_cleveland_dot_pre_calibrated_gemma}
    \end{subfigure}
    \begin{subfigure}[t]{0.32\textwidth}
        \centering
        \includegraphics[width=\textwidth]{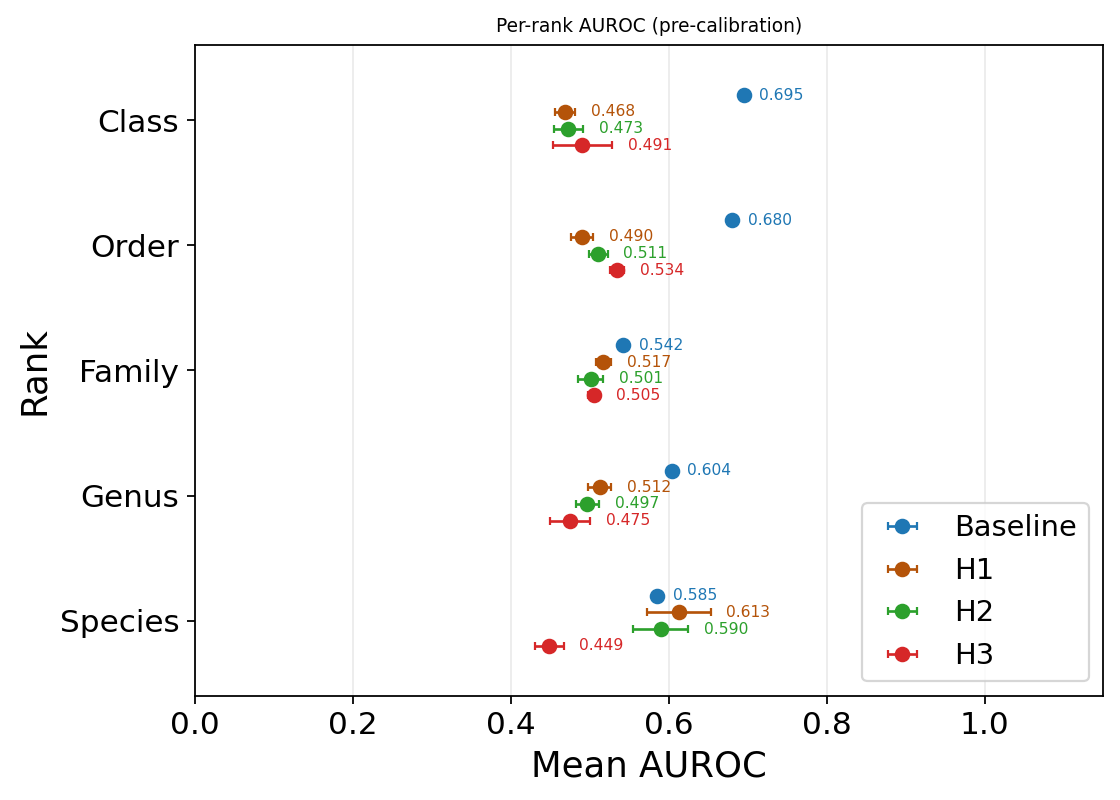}
        \caption{Qwen3 30B A3B Instruct.}
        \label{fig:per_rank_cleveland_dot_pre_calibrated_qwen3}
    \end{subfigure}
    \begin{subfigure}[t]{0.32\textwidth}
        \centering
        \includegraphics[width=\textwidth]{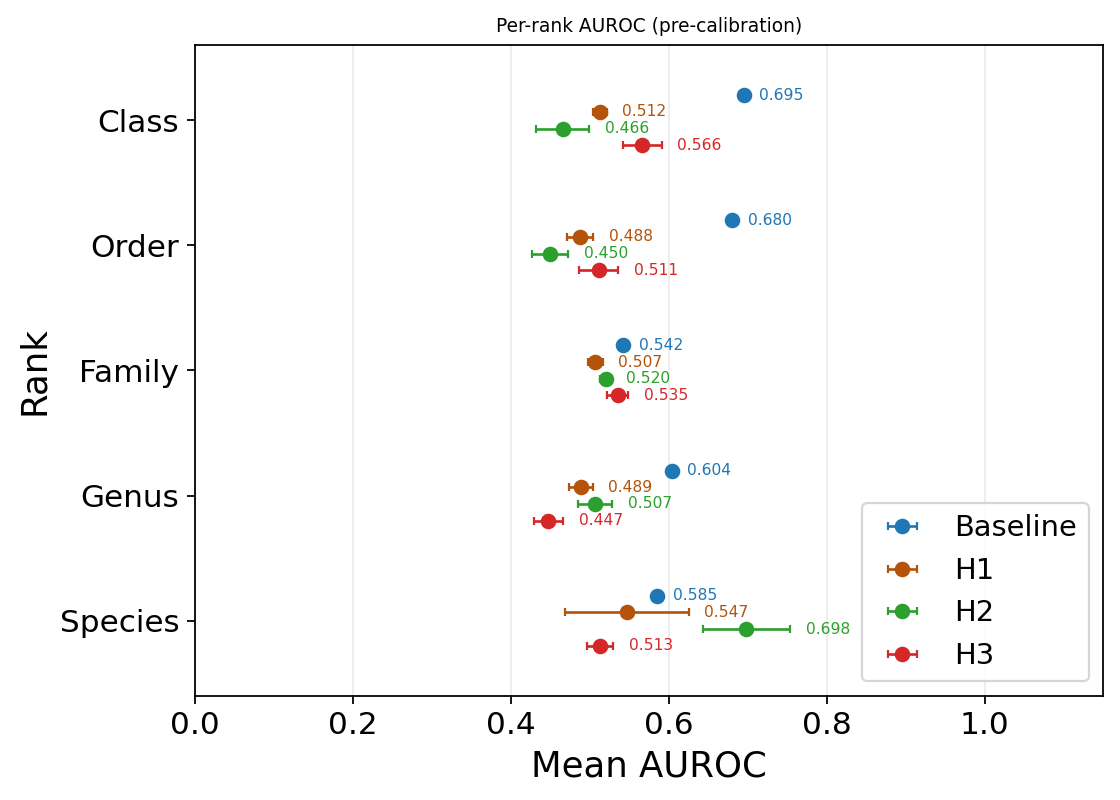}
        \caption{GPT-OSS 20B.}
        \label{fig:per_rank_cleveland_dot_pre_calibrated_gpt}
    \end{subfigure}
    \caption{Per-rank AUROC (pre-calibration). Markers show the mean over 5 random seeds, and horizontal bars indicate $\pm 1$ standard deviation.}
    \label{fig:per_rank_cleveland_dot_pre_calibrated_all}
\end{figure}

\newcommand{\ms}[2]{#1\,{\scriptsize(#2)}}
\newcommand{\msa}[3]{#1\textsuperscript{#3}\,{\scriptsize(#2)}}

\begin{table}[h]
\centering
% \scriptsize
% \setlength{\tabcolsep}{3.5pt}
% \renewcommand{\arraystretch}{0.9}
\caption{All-rank calibration metrics before vs.\ after temperature scaling~\citep{guo2017calibration}. Entries are mean (std across seeds), to three decimals. Baseline is model-invariant and shown once. \textcolor{ForestGreen}{\(\downarrow\)} indicates improvement (lower is better), and \textcolor{BrickRed}{\(\uparrow\)} indicates degradation. Bold marks the best (minimum) within each tool-LLM block for each metric column.}
\label{tab:all_rank_calibration_metrics_all_models}
\begin{tabular}{@{}llcccccc@{}}
\toprule
\multirow{2}{*}{Tool LLM} & \multirow{2}{*}{Method}
& \multicolumn{2}{c}{NLL} & \multicolumn{2}{c}{ECE} & \multicolumn{2}{c}{Brier} \\
\cmidrule(lr){3-4}\cmidrule(lr){5-6}\cmidrule(lr){7-8}
& & pre & cal & pre & cal & pre & cal \\
\midrule

None
& Baseline
& \ms{1.030}{0.000}
& \msa{0.706}{0.000}{\imp}
& \ms{0.114}{0.000}
& \msa{0.299}{0.000}{\degd}
& \ms{0.408}{0.000}
& \msa{0.256}{0.000}{\imp}
\\
\midrule

\multirow{3}{*}{Gemma 7B}
& H1
& \ms{0.488}{0.055}
& \msa{0.465}{0.039}{\imp}
& \ms{\textbf{0.057}}{0.018}
& \msa{0.064}{0.021}{\degd}
& \ms{0.150}{0.013}
& \msa{0.148}{0.011}{\imp}
\\
& H2
& \ms{0.470}{0.040}
& \msa{0.435}{0.026}{\imp}
& \ms{0.081}{0.023}
& \msa{\textbf{0.035}}{0.009}{\imp}
& \ms{0.151}{0.015}
& \msa{0.139}{0.011}{\imp}
\\
& H3
& \ms{\textbf{0.403}}{0.017}
& \msa{\textbf{0.411}}{0.018}{\degd}
& \ms{0.057}{0.020}
& \msa{0.059}{0.006}{\degd}
& \ms{\textbf{0.128}}{0.008}
& \msa{\textbf{0.127}}{0.006}{\imp}
\\
\midrule

\multirow{3}{*}{GPT-OSS 20B}
& H1
& \ms{0.437}{0.048}
& \msa{0.427}{0.038}{\imp}
& \ms{0.045}{0.024}
& \msa{0.037}{0.019}{\imp}
& \ms{0.144}{0.020}
& \msa{0.139}{0.015}{\imp}
\\
& H2
& \ms{\textbf{0.403}}{0.022}
& \msa{\textbf{0.400}}{0.018}{\imp}
& \ms{\textbf{0.038}}{0.016}
& \msa{\textbf{0.031}}{0.009}{\imp}
& \ms{\textbf{0.128}}{0.009}
& \msa{0.127}{0.006}{\imp}
\\
& H3
& \ms{0.411}{0.027}
& \msa{0.415}{0.008}{\degd}
& \ms{0.060}{0.023}
& \msa{0.055}{0.005}{\imp}
& \ms{0.130}{0.010}
& \msa{\textbf{0.125}}{0.003}{\imp}
\\
\midrule

\multirow{3}{*}{\makecell[l]{Qwen3 30B\\A3B Instruct}}
& H1
& \ms{\textbf{0.429}}{0.028}
& \msa{0.423}{0.028}{\imp}
& \ms{\textbf{0.040}}{0.016}
& \msa{\textbf{0.033}}{0.011}{\imp}
& \ms{\textbf{0.139}}{0.012}
& \msa{0.137}{0.011}{\imp}
\\
& H2
& \ms{0.446}{0.049}
& \msa{0.429}{0.036}{\imp}
& \ms{0.067}{0.044}
& \msa{0.041}{0.019}{\imp}
& \ms{0.147}{0.020}
& \msa{0.139}{0.014}{\imp}
\\
& H3
& \ms{0.447}{0.015}
& \msa{\textbf{0.410}}{0.030}{\imp}
& \ms{0.092}{0.026}
& \msa{0.053}{0.008}{\imp}
& \ms{0.145}{0.006}
& \msa{\textbf{0.129}}{0.010}{\imp}
\\
\bottomrule
\end{tabular}
\end{table}

\end{document}